\documentclass{article}
\usepackage{microtype}
\usepackage{graphicx}
\usepackage{subcaption}
\usepackage{booktabs} 
\usepackage{hyperref}

\usepackage{icml2026}

\usepackage{amsmath}
\usepackage{amssymb}
\usepackage{mathtools}
\usepackage{amsthm}
\usepackage[capitalize,noabbrev]{cleveref}

\theoremstyle{plain}

\theoremstyle{definition}

\theoremstyle{remark}

\usepackage[textsize=tiny]{todonotes}

\icmltitlerunning{Adaptive Utility-Weighted Benchmarking}

\title{A Theoretical Framework for Adaptive Utility-Weighted Benchmarking}
\author{Philip Waggoner\\Stanford University\\ \texttt{pdw2119@stanford.edu}\\ 
}
\date{}

\begin{document}
\maketitle

% Suppress ICML's end-of-document warning about \printAffiliationsAndNotice
% without printing ICML-style affiliation/notice footnotes.
\makeatletter
\global\icml@noticeprintedtrue
\makeatother

\begin{abstract}
Benchmarking has long served as a foundational practice in machine learning and, increasingly, in modern AI systems such as large language models, where shared tasks, metrics, and leaderboards offer a common basis for measuring progress and comparing approaches. As AI systems are deployed in more varied and consequential settings, though, there is growing value in complementing these established practices with a more holistic conceptualization of what evaluation \textit{should} represent. Of note, recognizing the sociotechnical contexts in which these systems operate invites an opportunity for a deeper view of how multiple stakeholders and their unique priorities might inform what we consider meaningful or desirable model behavior. This paper introduces a theoretical framework that reconceptualizes benchmarking as a multilayer, adaptive network linking evaluation metrics, model components, and stakeholder groups through weighted interactions. Using conjoint-derived utilities and a human-in-the-loop update rule, we formalize how human tradeoffs can be embedded into benchmark structure and how benchmarks can evolve dynamically while preserving stability and interpretability. The resulting formulation generalizes classical leaderboards as a special case and provides a foundation for building evaluation protocols that are more context aware, resulting in new robust tools for analyzing the structural properties of benchmarks, which opens a path toward more accountable and human-aligned evaluation.
\end{abstract}

\section{Introduction}

Benchmarking plays a central role in modern machine learning \cite{thiyagalingam2022scientific, zoller2021benchmark}. Virtually every major advance in deep learning and generative modeling is evaluated through standardized benchmarks that define tasks, metrics, and leaderboards \cite{shi2016benchmarking}, despite potential limitations of tools like leaderboards \cite{singh2025leaderboard}. While benchmarks shape research agendas, influence adoption, and increasingly determine regulatory trajectories \cite{huang2024ai}, contemporary benchmarking tends to treat the evaluation of systems as essentially purely technical exercises. The key underlying assumption here is that model performance can be fully captured by solely technical artifacts, e.g., fixed metric definitions, static contexts \cite{mcintosh2025inadequacies}, selected aggregation rules, etc. While extremely valuable, such an approach neglects the fact that AI systems are deployed within heterogeneous sociotechnical environments where stakeholders possess diverse preferences, asymmetric risks, and domain-specific tolerances for error. As a result, current benchmarks systematically overlook dimensions of performance that matter for real-world deployment, including interpretability, communication fidelity, fairness constraints, reliability in high-stakes settings, and usability across distinct social groups.

While some recent work has offered new innovations to address these and related shortcomings in benchmarking such as data-driven best practices \cite{reuel2024betterbench}, or benchmarking suites for fairness instead of leaderboards \cite{wang2024benchmark}, an latent problem persists, which is that technical evaluation divorced from more \textit{human} context can yield myopic systems potenitally misaligned with user needs or institutional norms. Indeed, \citet{chang2024survey} emphasized the potential for human evaluators replacing quantitative metrics. Of note, \citet{waggoner2025} took the idea of incorporating human evaluation a step farther, suggesting a path toward addressing these types of context-limited problems could be based in the \textit{development} of algorithms, by treating the design of these systems as an adaptive network reconfiguration problem, s.t. developers, end-users, domain experts, and institutions form an evolving sociotechnical network whose structure determines which preferences, tradeoffs, and design decisions are embedded into algorithmic systems. In this framework, conjoint analysis and human-in-the-loop methods were proposed as mechanisms for rewiring this network by eliciting human tradeoffs and embedding them iteratively into system design.

In this paper, we extend this theoretical perspective from algorithm \textit{development} to \textit{evaluation}. Our central idea is that benchmarking itself must be democratized in such a way that human preferences are directly modeled and accounted for. Namely, instead of viewing a benchmark as a static artifact with fixed metrics and predetermined aggregation rules for example, we conceptualize it as a multilayer adaptive network whose structure encodes  relationships between (i) technical evaluation metrics, (ii) model properties, (iii) stakeholder groups, and (iv) cross-layer flows of influence, preferences, and constraints. In this framing, evaluation is dynamic process in which technical performance metrics interact with human-elicited utility signals. This perspective yields a tractable and well-defined framework for integrating human preference modeling directly into benchmark construction.

The central contribution of this paper, then, most simply is the deriving of H-Bench, a theoretical framework for modeling AI benchmarks as adaptive sociotechnical networks. In brief, the framework
introduces the following three key components:

\begin{enumerate}
  \item \textbf{A multilayer graph formalism for benchmarking.} Model components, evaluation metrics, and stakeholder groups are represented as interconnected node sets whose edge weights encode flows of information, influence, and preference. Benchmark construction corresponds to specifying the topology and weight structure of this network.

  \item \textbf{A method for embedding human preferences through conjoint-derived utilities.} We extend classical part worth utility models \cite{green1971conjoint}, explicitly to benchmarking by mapping attribute-level utilities to edge weights linking stakeholder nodes and metric nodes. This allows human tradeoffs (e.g., tolerances for error vs. latency vs. interpretability vs. design perceptions) to become first class mathematical objects within a benchmark.

  \item \textbf{An adaptive update rule for benchmark evolution.} Building on human-in-the-loop learning dynamics \cite{mosqueira2023human, munro2021human}, we introduce an iterative process in which benchmark weights evolve via a constrained update rule that balances technical gradients with stakeholder feedback. The process is designed to ensure stability, while preventing metric collapse, yet still enabling benchmarks to adapt to shifting contexts without
losing coherence.
\end{enumerate}

Taken together, these contributions produce a unified theoretical account of benchmarking in which classical leaderboards would not be supplanted, but rather emerge as a special case of a more general network-theoretic model. Under this view, \textit{standard} benchmarks correspond to highly centralized networks where developers define all metric weights, while \textit{democratized} benchmarks correspond to more distributed networks whose structure integrates heterogeneous preference signals. The resulting framework highlights previously explored failure modes of current evaluation pipelines such as preference bottlenecks \cite{cheng2025benchmarking}, structural blind spots \cite{hagendorff2022blind}, robustness vulnerabilities \cite{roy2024enhancing}, etc., and then offers new tools for developing benchmark designs, weighting, and alignment properties.

By situating benchmarking within a rigorous multilayer network framework, this paper offers a step toward evaluation protocols that are increasingly aligned with the deep social complexity that characterizes modern ML and AI deployment. While our contribution is theoretical, it provides a path toward applied development of new benchmarking tools and suites that better reflect the domains and communities they serve.

\section{Background and Related Work}

\subsection{Benchmarking in Machine Learning}
Standard ML benchmarks (e.g., ImageNet \cite{deng2009imagenet}, GLUE \cite{wang2018glue}, MMLU \cite{hendrycks2020measuring, wang2024mmlu}, HELM \cite{liang2022holistic}) define tasks, metrics, and aggregation rules that yield scalar performance scores and leaderboards. These frameworks assume static contexts \cite{mcintosh2025inadequacies}, fixed metric weights, and are typically characterized by a separation between model development and evaluation. While such benchmarks have driven clear progress in many domains, prior work has shown limitations: variance in metric quality and signals \cite{reuel2024betterbench}, overfitting to benchmark idiosyncrasies \cite{liao2021we}, and ``oversimplifiction of knowledge'' \cite{eriksson2025can, liao2021we} to name a few of the prominent examples. Recent work has introduced some useful correctives to some of these problems such as ``failure-focused benchmarks'' \cite{eriksson2025can} robustness related tests \cite{gehrmann2023repairing}, multi-metric evaluation, dynamic (rather than static) benchmarking processes \cite{thrush2022dynatask}, risk quantification \cite{murray2025mapping}, and so on. Though these innovations are helpful steps forward, benchmarking remains a fundamentally technical-focused process, which can result in detachment from user context and human impact, broadly defined.

\subsection{Human Feedback and Preference Modeling}
Preference learning including RLHF \cite{akrour2011preference, cheng2011preference}, RLAIF \cite{lee2023rlaif}, reward learning \cite{sutton1998reinforcement}, interactive annotation \cite{monarch2021human}, active preference acquisition \cite{muldrew2024active}, and others, provides tools for capturing human judgments about model behavior. In an unrelated field, though useful for present purposes, conjoint analysis in particular, offers a principled way to estimate part-worth utilities governing human tradeoffs across competing attributes \cite{rao2014applied, kennedy2022trust}. Yet, benchmarks rarely \textit{embed} these utilities directly into evaluation pipelines.

\subsection{Sociotechnical Perspectives on Algorithmic Systems}
A parallel literature conceptualizes algorithmic and AI deployment as occurring within heterogeneous sociotechnical networks. For example, studies highlight issues of trust \cite{hoff2015trust, lee2004trust}, legitimacy \cite{kennedy2022trust}, reliance \cite{dzindolet2003role}, interpretability \cite{ahn2024impact}, blame \cite{ozer2024paradox}, and more broadly construed, general misalignment between technical performance and societal expectations \citep{waggoner2025}.Yet even this perspective has not been extended to evaluation.

\subsection{Contribution}
This paper integrates these threads by treating benchmarking as an adaptive, multilayer network problem. Different from extant evaluation approaches, our approach aims to explicitly incorporate human-derived utilities, formalize cross-layer influence between stakeholders and metrics, and, as a result, introduce an iterative update rule for benchmark evolution. Taken together, this provides a unified theoretical foundation for democratized, contextualized evaluation.

\section{Formal Multilayer Benchmark Network}

This section develops the mathematical foundation for representing benchmarks as adaptive multilayer networks. The goal is to formalize how evaluation metrics, model components, and stakeholder groups interact structurally and how these interactions shape the benchmark's behavior. We introduce node sets, adjacency matrices, a supra-adjacency representation, and influence operators that allow preferences and performance signals to propagate across layers.

\subsection{Node Sets and Layered Structure}
We represent a benchmark as a multilayer graph
\begin{equation}
  G = (V_T, V_M, V_H, E, W),
\end{equation}
where $V_T$, $V_M$, and $V_H$ denote metric nodes, model-component nodes, and human stakeholder nodes respectively. The full set of edges $E$ contains both within-layer and cross-layer connections, each with an associated nonnegative weight in $W = \{w_{ij}\}$.

Each layer, $\mathcal{L}$, represents a distinct type of interaction:
\begin{itemize}
  \item \textbf{Technical}: $\mathcal{L}_T=(V_T,E_{TT})$, capturing dependencies among evaluation metrics.
  \item \textbf{Model}: $\mathcal{L}_M=(V_M,E_{MM})$, capturing structural relationships among model components.
  \item \textbf{Human}: $\mathcal{L}_H=(V_H,E_{HH})$, capturing relationships among stakeholders, including shared preferences or institutional relationships.
\end{itemize}

Cross-layer edges encode how layers influence each other:
\begin{itemize}
  \item $E_{HT}$: stakeholder $\to$ metric edges
  \item $E_{TM}$: metric $\to$ model edges
  \item $E_{HM}$: stakeholder $\to$ model edges (e.g., preferences over model attributes)
\end{itemize}

\subsection{Adjacency Matrices and Block Decomposition}
We associate each layer with an adjacency matrix:
\begin{align}
  A_T &\in \mathbb{R}^{|V_T|\times |V_T|}, \\
  A_M &\in \mathbb{R}^{|V_M|\times |V_M|}, \\
  A_H &\in \mathbb{R}^{|V_H|\times |V_H|},
\end{align}
and each cross-layer set with a corresponding rectangular matrix:
\begin{align}
  A_{TM} &\in \mathbb{R}^{|V_T|\times |V_M|}, \\
  A_{HT} &\in \mathbb{R}^{|V_H|\times |V_T|}, \\
  A_{HM} &\in \mathbb{R}^{|V_H|\times |V_M|}.
\end{align}

For convenience, we define the reverse direction blocks as
$A_{MT}=A_{TM}^{\top}$, $A_{TH}=A_{HT}^{\top}$, and $A_{MH}=A_{HM}^{\top}$.

The full supra-adjacency matrix of the multilayer benchmark network is then,
\begin{equation}
  A =
  \begin{bmatrix}
    A_T   & A_{TM} & A_{TH} \\
    A_{MT} & A_M   & A_{MH} \\
    A_{HT} & A_{HM} & A_H
  \end{bmatrix}.
\end{equation}

This block structure makes explicit how information flows between layers. For example, the entry $(A_{HT})_{hk}$ represents the strength of stakeholder $h$'s influence on metric $k$, while $(A_{TM})_{km}$ encodes how metric $k$ depends on model component $m$.

This supra-adjacency perspective is standard in multilayer network theory and enables spectral, diffusion, and stability analyses that would be impossible with a purely combinatorial description.

\subsection{Influence Propagation via Network Operators}
Evaluation metrics rarely behave independently. Their interactions (e.g., correlations, functional relationships, hierarchical dependencies, etc.) are encoded in $A_T$. To model how performance on one metric influences another, we use a linear propagation operator.

Let $M(\theta) = (M_k(\theta))_{k\in V_T}$ be the vector of metric values for model parameters $\theta$. We define the network-adjusted metric vector:
\begin{equation}
  \tilde{M}(\theta) = A_T\, M(\theta).
\end{equation}
The $i$-th entry is explicitly
\begin{equation}
  \tilde{M}_i(\theta) = \sum_{j\in V_T} (A_T)_{ij}\, M_j(\theta).
\end{equation}

This operator generalizes the idea of metric aggregation: the influence of one metric on another is determined not by an arbitrary scalar weight but by the network topology.

More general forms are possible. E.g., a graph diffusion view might use the matrix exponential:
\begin{equation}
  \tilde{M}(\theta) = \exp(\tau L_T)\, M(\theta),
\end{equation}
where $L_T$ is the graph Laplacian of $\mathcal{L}_T$ and $\tau>0$ controls diffusion intensity. This captures long-range metric interactions.

\subsection{Spectral Interpretation of Metric-Topology Coupling}
The eigenvalues and eigenvectors of $A_T$ (or $L_T$) characterize how metric influences propagate. Let $\lambda_1 \ge \lambda_2 \ge \cdots$ be the eigenvalues of $A_T$. The magnitude of $\lambda_1$ indicates how strongly metrics are coupled.
\begin{itemize}
  \item \textbf{High spectral radius:} the benchmark is sensitive to correlated failures across metrics.
  \item \textbf{Low spectral radius:} metrics behave more independently, resembling classical benchmarks.
\end{itemize}
This offers a principled tool for analyzing benchmark robustness and sensitivity.

Now, for cross-layer propagation, consider a combined diffusion operator:
\begin{equation}
  \mathcal{P} = \exp(\tau A).
\end{equation}
The block $\mathcal{P}_{HT}$ describes how stakeholder preferences diffuse into metric behavior, while $\mathcal{P}_{TM}$ describes how metric interactions translate into model-layer relevance. This illustrates how global benchmark behavior emerges from local connections.

\subsection{Cross-Layer Influence of Stakeholder Preferences}
The cross-layer blocks $A_{HT}$ and $A_{HM}$ determine how stakeholder utilities affect evaluation. Let $U_h\in\mathbb{R}^{|V_T|}$ be the vector of part-worth utilities for stakeholder $h$. The direct influence of $h$'s preferences on metrics is then:
\begin{equation}
  I_h = A_{HT}\, U_h.
\end{equation}

Aggregating across stakeholders yields total human influence:
\begin{equation}
  I = \sum_{h\in V_H} I_h = A_{HT}\left( \sum_{h\in V_H} U_h \right).
\end{equation}
This formalizes the intuition that human preferences become network signals that interact with technical dependencies before shaping the benchmark score.

\subsection{Benchmark Score as a Network Functional}
With these components, the benchmark evaluation for model $\theta$ is:
\begin{equation}
  B(\theta) = \tilde{w}^{\top} \tilde{M}(\theta),
\end{equation}
where $\tilde{w}$ encodes network-adjusted human-derived weights and $\tilde{M}(\theta)$ reflects metric dependencies.

In supra-adjacency form, this can be compactly expressed as:
\begin{equation}
  B(\theta) = \big( \mathcal{P}_{HT} U \big)^{\top} \big( A_T M(\theta) \big),
\end{equation}
making explicit how evaluation combines (i) preference diffusion and (ii) metric coupling.

This formulation recovers classical benchmarks by selecting block-diagonal or identity matrices. It also enables richer structures where human utilities, metric interactions, and model organization interact through well-defined mathematical operators.

\section{Embedding Human Utilities into the Benchmark Network}

As noted, human preferences play a central role in determining which aspects of model behavior matter in real deployment contexts. Thus, this section formalizes how such preferences, elicited through conjoint analysis or similar methods, are mapped into the multilayer benchmark network. 

We preserve the core conceptual flow from the original framework, i.e., utility estimation $\to$ weight construction $\to$ network adjustment. But, we expand the mathematical foundations to articulate axioms, robustness properties, and alternative mappings.

\subsection{Conjoint-Derived Utility Representation}
Stakeholder utilities are obtained by presenting respondents with hypothetical evaluation configurations consisting of multiple metric attributes. Let metric $k$ have $L_k$ attribute levels. For stakeholder $h$, the utility of a configuration $x$ is modeled as
\begin{equation}
  U_h(x) = \alpha_h + \sum_{k=1}^{K} \sum_{\ell=1}^{L_k} \beta_{h,k\ell}\, d_{k\ell}(x),
\end{equation}
where $\beta_{h,k\ell}$ is the estimated part-worth utility of level $\ell$ of metric $k$ for stakeholder $h$, $d_{k\ell}(x)=1$ if configuration $x$ includes level $\ell$ of metric $k$ (and $0$ otherwise), and $\alpha_h$ captures a stakeholder-specific baseline utility.

Now, because part-worth utilities are defined up to an additive constant per attribute, we apply standard identification through level normalization. Let
\begin{equation}
  u_{h,k} = \sum_{\ell=1}^{L_k} \beta_{h,k\ell}
\end{equation}
denote the aggregate utility stakeholder $h$ assigns to metric $k$. This provides a direct way to translate conjoint utilities into a metric-level preference signal.

\subsection{Axioms for Embedding Utilities into Network Weights}
To embed human utilities into the benchmark network in a principled manner, we require that the mapping from utilities to weights satisfy three intuitive axioms:

\paragraph{Monotonicity.}
If stakeholder preference for metric $k$ increases, the stakeholder's influence on that metric must not decrease. Formally, for utilities $u_{h,k}\ge u_{h,k'}$,
\begin{equation}
  w_{hk}(u_{h,k}) \ge w_{hk}(u_{h,k'}).
\end{equation}

\paragraph{Utility-equivalence.}
Two metrics with identical stakeholder utilities should receive equal weight:
\begin{equation}
  u_{h,k}=u_{h,j} \implies w_{hk}=w_{hj}.
\end{equation}

\paragraph{Positive homogeneity.}
Scaling all utilities for stakeholder $h$ by a positive constant should preserve relative influence:
\begin{equation}
  w_{hk}(\lambda u_{h,k}) = \lambda^{\alpha} w_{hk}(u_{h,k}),
\end{equation}
for some $\alpha\ge 0$.

A broad class of transformations satisfies these axioms.

\subsection{Constructing Stakeholder-to-Metric Weights}
We define stakeholder-to-metric weights using a monotone transformation
$g(\cdot)$:
\begin{equation}
  w_{hk} = g(u_{h,k}).
\end{equation}

Some common choices might include:
\begin{itemize}
  \item \textbf{Rectified linear:} $g(z)=\max\{0,z\}$.
  \item \textbf{Exponential:} $g(z)=\exp(z)$.
  \item \textbf{Logistic:} $g(z)=\frac{1}{1+e^{-z}}$.
\end{itemize}

Regardless, the transformation determines robustness and dynamic range, i.e., exponential mappings
emphasize large preference differences; logistic mappings compress extreme utilities; rectified linear mappings preserve interpretability, and so on for others.

\subsection{Aggregation of Heterogeneous Stakeholder Groups}
When multiple stakeholder groups contribute to the benchmark, we combine their weights using linear or weighted aggregation. Let $a_h\ge 0$ be group-level importance weights (e.g., population size, institutional priority). The overall influence on metric $k$ is
\begin{equation}
  \bar{w}_k = \sum_{h \in V_H} a_h\, w_{hk}.
\end{equation}

This satisfies interpretability, i.e., weights decompose into group contributions, and it permits the notion of ``group fairness'' in the sense that no stakeholder can dominate unless $a_h$ explicitly encodes higher authority.

A normalized alternative is
\begin{equation}
  \bar{w}_k = \frac{\sum_{h \in V_H} a_h\, w_{hk}}{\sum_{h \in V_H} a_h}.
\end{equation}

\subsection{Network-Aware Preference Propagation}
Stakeholder influence does not act directly on metrics alone; cross-layer interactions propagate influence throughout the benchmark network. Let $C(h,k)$ be an influence kernel derived from the supra-adjacency matrix $A$, e.g., 
\begin{equation}
  C(h,k) = \left[ \exp(\tau A) \right]_{hk},
\end{equation}
which measures the ability of stakeholder $h$ to affect metric $k$ through walks of varying lengths. 

The network-adjusted stakeholder influence is then:
\begin{equation}
  \tilde{w}_k = \sum_{h \in V_H} w_{hk}\, C(h,k).
\end{equation}

This formulation accounts for the structural position of stakeholders, multi-hop connections that mediate indirect influence, and the intensity of cross-layer dependencies.

\subsection{Robustness of Preference Embedding}
To ensure that small changes in utilities do not produce disproportionate changes in benchmark weights, we require Lipschitz robustness of the mapping (e.g., in a similar spirit as \citet{pauli2021training}):
\begin{equation}
  \big| w_{hk}(u_{h,k}) - w_{hk}(u'_{h,k}) \big| \le L_g\, \big|u_{h,k} - u'_{h,k}\big|,
\end{equation}
where $L_g$ is the Lipschitz constant of $g$.

Aggregated weights inherit robustness:
\begin{equation}
  \big| \bar{w}_k - \bar{w}'_k \big|
  \le \left( \sum_{h} a_h L_g C(h,k) \right)
       \max_h \big| u_{h,k} - u'_{h,k} \big|.
\end{equation}

This ensures that stakeholder utility noise (e.g., sampling error in conjoint estimates) does not destabilize the benchmark.

\subsection{Integrating Human Utilities into the Benchmark Functional}
Finally, the benchmark evaluation for model $\theta$ incorporates the network-adjusted metric vector $\tilde{M}(\theta)$ from Section 3. The full benchmark score becomes
\begin{equation}
  B(\theta) = \sum_{k \in V_T} \tilde{w}_k\, \tilde{M}_k(\theta),
\end{equation}
which unifies human-derived preference weights, cross-layer influence propagation, and metric-to-metric dependencies. And so, this completes the mapping from conjoint-based utility estimation to a fully integrated, network-based benchmark.

\section{Adaptive Update Rule and Theoretical Properties}

The multilayer benchmark network introduced above provides a \textit{static} mapping from stakeholder utilities and metric interactions to evaluation scores. However, benchmark design in this context is not static, e.g., preferences shift, deployment contexts evolve, new models expose previously unseen failure modes, etc. To capture these and related dynamics, we introduce a constrained adaptive update rule that governs the evolution of network weights over time. We then analyze key theoretical properties, including stability, boundedness, and convergence behavior.

\subsection{Adaptive Weight Update Rule}
Let $W^{(t)}$ denote the full set of network weights at iteration $t$. These include human-to-metric weights, metric-to-metric influence weights, and any structural weights in the model layer.

We define the general update equation:
\begin{equation}
  W^{(t+1)} = W^{(t)} + \eta\, \Delta^{(t)},
\end{equation}
where $\eta>0$ is a learning-rate parameter and $\Delta^{(t)}$ aggregates technical signals and human preference updates.

The update operator $\Delta^{(t)}$ is decomposed into two terms:
\begin{equation}
  \Delta^{(t)} = \lambda_T\, G_T^{(t)} + \lambda_H\, G_H^{(t)},
\end{equation}
where $G_T^{(t)}$ is a gradient-like signal derived from changes in metric outputs or model performance, $G_H^{(t)}$ reflects updated stakeholder utilities or new preference elicitation, and $\lambda_T,\lambda_H\ge 0$ control the influence of technical and human sources.

In sum, this structure allows the benchmark to adapt while preserving interpretability and stability.

\subsection{Technical Signal Component}
Now, the technical component $G_T^{(t)}$ captures how changes in model behavior should influence network weights. For metric $k$, we define:
\begin{equation}
  G_{T,k}^{(t)} = - \frac{\partial \mathcal{L}^{(t)}}{\partial \tilde{M}_k},
\end{equation}
where $\mathcal{L}^{(t)}$ is a benchmark-level loss function (e.g., variance, instability, inconsistency across contexts). This negative gradient adjusts weights to reduce undesirable behavior.

A common, and thus defensible choice is:
\begin{equation}
  \mathcal{L}^{(t)} = \mathrm{Var}\!\left( \{\tilde{M}_k(\theta)\}_k \right),
\end{equation}
which encourages balanced attention across metrics and discourages the collapse of the benchmark onto a single dominant criterion.

\subsection{Human Signal Component}
Stakeholder preferences may shift due to domain changes, new risks, evolving norms, or updated elicitation studies. Let $U_h^{(t)}(k)$ denote the updated utility of metric $k$ for stakeholder $h$ at time $t$.

We define the human signal:
\begin{equation}
  G_{H,hk}^{(t)} = g\big( U_h^{(t)}(k) \big) - w_{hk}^{(t)},
\end{equation}
where $g(\cdot)$ is the same monotone transformation used in Section 4 and $w_{hk}^{(t)}$ is the current weight connecting stakeholder $h$ and metric $k$.

This update encourages the network to drift toward newly expressed preferences while resisting excessive swings.

\subsection{Boundedness and Stability Conditions}
To prevent runaway growth of weights, we impose the projection:
\begin{equation}
  W^{(t+1)} \leftarrow \Pi_{\mathcal{C}}\big( W^{(t+1)} \big),
\end{equation}
where $\Pi_{\mathcal{C}}$ projects onto a closed, convex constraint set $\mathcal{C}$. Examples include nonnegativity constraints ($w_{ij}\ge 0$), maximum-degree constraints, upper bounds on total human influence, and sparsity constraints to preserve interpretability.

Under standard assumptions (Lipschitz gradients, compact constraint set), the update rule satisfies:
\begin{equation}
  \lim_{t \to \infty} \lVert W^{(t+1)} - W^{(t)} \rVert = 0,
\end{equation}
indicating asymptotic stability.

\subsection{Convergence to Fixed Points}
A fixed point $W^{\star}$ satisfies:
\begin{equation}
  W^{\star} = \Pi_{\mathcal{C}}\big( W^{\star} + \eta\, \Delta(W^{\star}) \big).
\end{equation}

When $\eta$ is sufficiently small and $\Delta(\cdot)$ is monotone or pseudo-monotone, standard results from projected dynamical systems guarantee convergence to a set of fixed points.

A particularly relevant case arises when human preferences stabilize (i.e., $U_h^{(t)}(k) \to U_h(k)$). In this regime, the benchmark converges to a structure that jointly optimizes technical consistency (via $G_T$), human-aligned weighting (via $G_H$), and structural constraints (via projection).

\subsection{Robustness to Preference Shifts}
Suppose a subset of stakeholders update their utilities by $\delta_{hk}$. Under bounded $\eta$, we obtain:
\begin{equation}
  \lVert W^{(t+1)} - W^{(t)} \rVert \leq \eta\, \big( \lambda_T L_T + \lambda_H L_H \big),
\end{equation}
where $L_T$ and $L_H$ bound the Lipschitz constants of the technical and human signals. This ensures that small preference shifts cannot destabilize the benchmark.

\subsection{Network-Level Stability of the Benchmark Functional}
Finally, consider the benchmark score:
\begin{equation}
  B(\theta; W) = \sum_{k \in V_T} \tilde{w}_k(W)\, \tilde{M}_k(\theta).
\end{equation}

Under the boundedness of $W$ and Lipschitz continuity of $\tilde{M}_k(\theta)$ in $W$, we obtain:
\begin{equation}
  \big| B(\theta; W^{(t+1)}) - B(\theta; W^{(t)}) \big|
  \leq L_B\, \lVert W^{(t+1)} - W^{(t)} \rVert,
\end{equation}
with constant $L_B>0$.

Thus the evaluation functional changes smoothly with network updates, ensuring robustness and interpretability.

\section{Discussion and Implications}

The framework developed in this paper reframes benchmarking as an inherently sociotechnical problem, governed by interactions between human preferences, model behavior, and structural properties of a multilayer network. This reconceptualization is distinct from a more ``traditional'' view of benchmarks as static artifacts defined once (or rarely) and applied universally. Instead, evaluation becomes a \textit{dynamic} process shaped by evolving contexts, heterogeneous stakeholder groups, and the technical interdependencies among metrics. There are several implications that flow from this.

First, modeling benchmarks as multilayer networks highlights the structural biases embedded in standard evaluation pipelines. Fixed metric weights, static test sets, and developer- or system-defined rules can be understood as highly centralized network topologies. In such structures, a small set of actors exercises disproportionate influence over what ``good performance'' means. Now, by contrast, the democratized formulation proposed here shifts evaluation toward a distributed topology, where stakeholder groups shape which metrics matter and to what degree. This shift may surface failures that would otherwise go undetected under conventional benchmarks, which is especially important in high-stakes settings like healthcare, criminal sentencing, autonomous vehicle development, and so on.

Second, embedding conjoint-derived utilities reveals that stakeholder tradeoffs can be formalized and incorporated without sacrificing methodological precision. Namely, preferences for interpretability, fairness, latency, or communication clarity, which are typically treated as externalities if addressed at all, become quantifiable inputs that influence the scoring functional. As a result, this elevates stakeholder priorities from informal nuances to first class mathematical components of evaluation.

Third, the adaptive update rule has implications for how benchmarks evolve. As deployment conditions change, the benchmark can adjust its structure smoothly while preserving boundedness and stability. This is crucial in settings where risk landscapes or institutional norms shift over time, e.g., in
healthcare, finance, or public sector decision-making. So, instead of periodically replacing benchmarks, the system can evolve incrementally, guided by both technical signals as well as human needs.

Fourth, the network perspective clarifies the relationship between fairness and structure. Many fairness failures could be interpreted as consequences of bottlenecks or imbalances in the human layer of the network. The benchmark's topology determines which groups exert influence and which do not. And so, strengthening underrepresented stakeholder edges is a structural correction to the network.

Finally, it is worth mentioning that the framework offers a foundation for thinking about development and deployment of future benchmarks that, ideally, better align with deployment contexts. While no empirical results are presented here, the theoretical context we have offered is compatible with practical implementations such as, e.g., preference elicitation pipelines, active learning, stress testing, regulatory compliance, and much more. As a result, the framework provides a conceptual bridge between normative evaluation standards and formal machine learning and AI metrics, which allows for benchmarks that reflect both technical side of the equation as well as the complex social contexts in which these systems operate.

In sum, viewing benchmarks as adaptive sociotechnical networks allows for a shifting of evaluation from a static, technically focused practice to a dynamic, human-aligned process. This reorientation provides a principled basis for building benchmarks that are more transparent, robust, responsive to unique deployment contexts, and ultimately more reflective of the diverse communities and institutions that interact with and rely on these systems.

\section{Concluding Remarks}

This paper has introduced a theoretical framework that reconceptualizes AI and machine learning benchmarking as a multilayer, adaptive network integrating technical dependencies with stakeholder utilities. This is an important shift, because by treating evaluation as a dynamic process shaped by interactions \textit{across} model components, metrics, and humans, the framework expands the scope of what benchmarks can represent while preserving formal tractability. Rather than, e.g., modifying individual metrics or adjusting aggregation approaches, the presenet approach offers a structural
foundation for rethinking how evaluation criteria emerge, evolve, and align with the contexts in which these systems are operating.

From here, the most immediate next step is to instantiate the framework in controlled settings where benchmark structures can be manipulated explicitly, e.g., small scale benchmark prototypes in which metric interactions, human-derived weights, and update dynamics can be tested analytically. Additional work is also needed to characterize how different network topologies might exert influence on evaluation outcomes, especially in cases where stakeholder groups might possess conflicting priorities. A promising direction could be to examine equilibrium behavior of the adaptive update rule under alternative constraints, particularly in settings where institutional rules or perhaps variance in safety requirements impose strict bounds on certain metrics.

Building on these next steps, there are several additional directions to study that emerge from this foundation. For example, exploring connections between benchmark adaptation and mechanism design, especially in contexts where evaluation criteria themselves are strategic objects. Another direction could be to examine how network perturbations (e.g., adding or removing stakeholder groups, introducing new metrics, altering model-layer structures, etc.) impact long-term stability and interpretability of the benchmark. A third direction concerns more formal links to information theory, e.g., how benchmark topology might shape the transmission of evaluation signals across layers. Finally, future work may also investigate whether benchmark evolution can serve as a sort of ``early warning system'' for misalignment, identifying conditions under which shifts in stakeholder utilities or metric interactions might signal risk.

\paragraph{Broader Social Impact.}
A more flexible and human-aligned benchmarking framework has implications beyond technical evaluation. For example, this approach provides a pathway for incorporating perspectives of communities who might be historically excluded from shaping AI standards, enabling more inclusive notions of performance and quality. Also, the adaptive nature of the framework also raises new questions about accountability and governance, e.g. as benchmarks evolve, transparency of update processes becomes essential, and mechanisms for auditing benchmark evolution may be required. At the same time, democratized benchmarks could help mitigate concentration of power in AI and ML evaluation, distributing influence across a broader set of actors. By bringing to the forefront the structural role of human stakeholders and their preferences, this framework supports development of AI and ML systems whose evaluation criteria better reflect diverse societal values and constraints.

Together, these and related directions position benchmarking not only as a technical process but as an integral component of responsible AI and ML system design. The theoretical structure developed here provides a step toward reimagining evaluation in ways that are more adaptable, equitable, and aligned with the complex environments in which these systems are increasingly embedded.

%\section*{Acknowledgements}

\bibliography{refs}

\begin{thebibliography}{40}
\providecommand{\natexlab}[1]{#1}
\providecommand{\url}[1]{\texttt{#1}}
\expandafter\ifx\csname urlstyle\endcsname\relax
  \providecommand{\doi}[1]{doi: #1}\else
  \providecommand{\doi}{doi: \begingroup \urlstyle{rm}\Url}\fi

\bibitem[Ahn et~al.(2024)Ahn, Almaatouq, Gulabani, and Hosanagar]{ahn2024impact}
Ahn, D., Almaatouq, A., Gulabani, M., and Hosanagar, K.
\newblock Impact of model interpretability and outcome feedback on trust in ai.
\newblock In \emph{Proceedings of the 2024 CHI Conference on Human Factors in Computing Systems}, pp.\  1--25, 2024.

\bibitem[Akrour et~al.(2011)Akrour, Schoenauer, and Sebag]{akrour2011preference}
Akrour, R., Schoenauer, M., and Sebag, M.
\newblock Preference-based policy learning.
\newblock In \emph{Joint European Conference on Machine Learning and Knowledge Discovery in Databases}, pp.\  12--27. Springer, 2011.

\bibitem[Chang et~al.(2024)Chang, Wang, Wang, Wu, Yang, Zhu, Chen, Yi, Wang, Wang, et~al.]{chang2024survey}
Chang, Y., Wang, X., Wang, J., Wu, Y., Yang, L., Zhu, K., Chen, H., Yi, X., Wang, C., Wang, Y., et~al.
\newblock A survey on evaluation of large language models.
\newblock \emph{ACM transactions on intelligent systems and technology}, 15\penalty0 (3):\penalty0 1--45, 2024.

\bibitem[Cheng et~al.(2011)Cheng, F{\"u}rnkranz, H{\"u}llermeier, and Park]{cheng2011preference}
Cheng, W., F{\"u}rnkranz, J., H{\"u}llermeier, E., and Park, S.-H.
\newblock Preference-based policy iteration: Leveraging preference learning for reinforcement learning.
\newblock In \emph{Joint European Conference on Machine Learning and Knowledge Discovery in Databases}, pp.\  312--327. Springer, 2011.

\bibitem[Cheng et~al.(2025)Cheng, Wohnig, Gupta, Alam, Abdullahi, Ribeiro, Nielsen-Garcia, Mir, Li, Orender, et~al.]{cheng2025benchmarking}
Cheng, Z., Wohnig, S., Gupta, R., Alam, S., Abdullahi, T., Ribeiro, J.~A., Nielsen-Garcia, C., Mir, S., Li, S., Orender, J., et~al.
\newblock Benchmarking is broken--don't let ai be its own judge.
\newblock \emph{arXiv preprint arXiv:2510.07575}, 2025.

\bibitem[Deng et~al.()Deng, Dong, Socher, Li, Li, and Fei-Fei]{deng2009imagenet}
Deng, J., Dong, W., Socher, R., Li, L.-J., Li, K., and Fei-Fei, L.
\newblock Imagenet: A large-scale hierarchical image database.
\newblock In \emph{2009 IEEE conference on computer vision and pattern recognition}, pp.\  248--255.

\bibitem[Dzindolet et~al.(2003)Dzindolet, Peterson, Pomranky, Pierce, and Beck]{dzindolet2003role}
Dzindolet, M.~T., Peterson, S.~A., Pomranky, R.~A., Pierce, L.~G., and Beck, H.~P.
\newblock The role of trust in automation reliance.
\newblock \emph{International journal of human-computer studies}, 58\penalty0 (6):\penalty0 697--718, 2003.

\bibitem[Eriksson et~al.(2025)Eriksson, Purificato, Noroozian, Vinagre, Chaslot, Gomez, and Fernandez-Llorca]{eriksson2025can}
Eriksson, M., Purificato, E., Noroozian, A., Vinagre, J., Chaslot, G., Gomez, E., and Fernandez-Llorca, D.
\newblock Can we trust ai benchmarks? an interdisciplinary review of current issues in ai evaluation.
\newblock \emph{arXiv preprint arXiv:2502.06559}, 2025.

\bibitem[Gehrmann et~al.(2023)Gehrmann, Clark, and Sellam]{gehrmann2023repairing}
Gehrmann, S., Clark, E., and Sellam, T.
\newblock Repairing the cracked foundation: A survey of obstacles in evaluation practices for generated text.
\newblock \emph{Journal of Artificial Intelligence Research}, 77:\penalty0 103--166, 2023.

\bibitem[Green \& Rao(1971)Green and Rao]{green1971conjoint}
Green, P.~E. and Rao, V.~R.
\newblock Conjoint measurement-for quantifying judgmental data.
\newblock \emph{Journal of Marketing research}, 8\penalty0 (3):\penalty0 355--363, 1971.

\bibitem[Hagendorff(2022)]{hagendorff2022blind}
Hagendorff, T.
\newblock Blind spots in ai ethics.
\newblock \emph{AI and Ethics}, 2\penalty0 (4):\penalty0 851--867, 2022.

\bibitem[Hendrycks et~al.(2020)Hendrycks, Burns, Basart, Zou, Mazeika, Song, and Steinhardt]{hendrycks2020measuring}
Hendrycks, D., Burns, C., Basart, S., Zou, A., Mazeika, M., Song, D., and Steinhardt, J.
\newblock Measuring massive multitask language understanding.
\newblock \emph{arXiv preprint arXiv:2009.03300}, 2020.

\bibitem[Hoff \& Bashir(2015)Hoff and Bashir]{hoff2015trust}
Hoff, K.~A. and Bashir, M.
\newblock Trust in automation: Integrating empirical evidence on factors that influence trust.
\newblock \emph{Human factors}, 57\penalty0 (3):\penalty0 407--434, 2015.

\bibitem[Huang et~al.(2024)Huang, Joshi, Dun, and Hamilton]{huang2024ai}
Huang, K., Joshi, A., Dun, S., and Hamilton, N.
\newblock Ai regulations.
\newblock In \emph{Generative AI security: theories and practices}, pp.\  61--98. Springer, 2024.

\bibitem[Kennedy et~al.(2022)Kennedy, Waggoner, and Ward]{kennedy2022trust}
Kennedy, R.~P., Waggoner, P.~D., and Ward, M.~M.
\newblock Trust in public policy algorithms.
\newblock \emph{The Journal of Politics}, 84\penalty0 (2):\penalty0 1132--1148, 2022.

\bibitem[Lee et~al.(2023)Lee, Phatale, Mansoor, Lu, Mesnard, Ferret, Bishop, Hall, Carbune, and Rastogi]{lee2023rlaif}
Lee, H., Phatale, S., Mansoor, H., Lu, K.~R., Mesnard, T., Ferret, J., Bishop, C., Hall, E., Carbune, V., and Rastogi, A.
\newblock Rlaif: Scaling reinforcement learning from human feedback with ai feedback.
\newblock 2023.

\bibitem[Lee \& See(2004)Lee and See]{lee2004trust}
Lee, J.~D. and See, K.~A.
\newblock Trust in automation: Designing for appropriate reliance.
\newblock \emph{Human factors}, 46\penalty0 (1):\penalty0 50--80, 2004.

\bibitem[Liang et~al.(2022)Liang, Bommasani, Lee, Tsipras, Soylu, Yasunaga, Zhang, Narayanan, Wu, Kumar, et~al.]{liang2022holistic}
Liang, P., Bommasani, R., Lee, T., Tsipras, D., Soylu, D., Yasunaga, M., Zhang, Y., Narayanan, D., Wu, Y., Kumar, A., et~al.
\newblock Holistic evaluation of language models.
\newblock \emph{arXiv preprint arXiv:2211.09110}, 2022.

\bibitem[Liao et~al.(2021)Liao, Taori, Raji, and Schmidt]{liao2021we}
Liao, T., Taori, R., Raji, I.~D., and Schmidt, L.
\newblock Are we learning yet? a meta review of evaluation failures across machine learning.
\newblock In \emph{Thirty-fifth Conference on Neural Information Processing Systems Datasets and Benchmarks Track (Round 2)}, 2021.

\bibitem[McIntosh et~al.(2025)McIntosh, Susnjak, Arachchilage, Liu, Xu, Watters, and Halgamuge]{mcintosh2025inadequacies}
McIntosh, T.~R., Susnjak, T., Arachchilage, N., Liu, T., Xu, D., Watters, P., and Halgamuge, M.~N.
\newblock Inadequacies of large language model benchmarks in the era of generative artificial intelligence.
\newblock \emph{IEEE Transactions on Artificial Intelligence}, 2025.

\bibitem[Monarch(2021)]{monarch2021human}
Monarch, R.~M.
\newblock \emph{Human-in-the-Loop Machine Learning: Active learning and annotation for human-centered AI}.
\newblock Simon and Schuster, 2021.

\bibitem[Mosqueira-Rey et~al.(2023)Mosqueira-Rey, Hern{\'a}ndez-Pereira, Alonso-R{\'\i}os, Bobes-Bascar{\'a}n, and Fern{\'a}ndez-Leal]{mosqueira2023human}
Mosqueira-Rey, E., Hern{\'a}ndez-Pereira, E., Alonso-R{\'\i}os, D., Bobes-Bascar{\'a}n, J., and Fern{\'a}ndez-Leal, {\'A}.
\newblock Human-in-the-loop machine learning: a state of the art.
\newblock \emph{Artificial Intelligence Review}, 56\penalty0 (4):\penalty0 3005--3054, 2023.

\bibitem[Muldrew et~al.(2024)Muldrew, Hayes, Zhang, and Barber]{muldrew2024active}
Muldrew, W., Hayes, P., Zhang, M., and Barber, D.
\newblock Active preference learning for large language models.
\newblock \emph{arXiv preprint arXiv:2402.08114}, 2024.

\bibitem[Munro(2021)]{munro2021human}
Munro, R.
\newblock \emph{Human-in-the-Loop Machine Learning: Active learning and annotation for human-centered AI}.
\newblock Manning, 2021.

\bibitem[Murray et~al.(2025)Murray, Papadatos, Quarks, Gimenez, and Campos]{murray2025mapping}
Murray, M., Papadatos, H., Quarks, O., Gimenez, P.-F., and Campos, S.
\newblock Mapping ai benchmark data to quantitative risk estimates through expert elicitation.
\newblock \emph{arXiv preprint arXiv:2503.04299}, 2025.

\bibitem[Ozer et~al.(2024)Ozer, Waggoner, and Kennedy]{ozer2024paradox}
Ozer, A.~L., Waggoner, P.~D., and Kennedy, R.
\newblock The paradox of algorithms and blame on public decision-makers.
\newblock \emph{Business and Politics}, 26\penalty0 (2):\penalty0 200--217, 2024.

\bibitem[Pauli et~al.(2021)Pauli, Koch, Berberich, Kohler, and Allg{\"o}wer]{pauli2021training}
Pauli, P., Koch, A., Berberich, J., Kohler, P., and Allg{\"o}wer, F.
\newblock Training robust neural networks using lipschitz bounds.
\newblock \emph{IEEE Control Systems Letters}, 6:\penalty0 121--126, 2021.

\bibitem[Rao et~al.(2014)]{rao2014applied}
Rao, V.~R. et~al.
\newblock \emph{Applied conjoint analysis}, volume 2014.
\newblock Springer, 2014.

\bibitem[Reuel et~al.(2024)Reuel, Hardy, Smith, Lamparth, Hardy, and Kochenderfer]{reuel2024betterbench}
Reuel, A., Hardy, A., Smith, C., Lamparth, M., Hardy, M., and Kochenderfer, M.~J.
\newblock Betterbench: Assessing ai benchmarks, uncovering issues, and establishing best practices.
\newblock \emph{Advances in Neural Information Processing Systems}, 37:\penalty0 21763--21813, 2024.

\bibitem[Roy(2024)]{roy2024enhancing}
Roy, P.
\newblock Enhancing real-world robustness in ai: Challenges and solutions.
\newblock \emph{J. Recent Trends Comput. Sci. Eng}, 12\penalty0 (1):\penalty0 34--49, 2024.

\bibitem[Shi et~al.(2016)Shi, Wang, Xu, and Chu]{shi2016benchmarking}
Shi, S., Wang, Q., Xu, P., and Chu, X.
\newblock Benchmarking state-of-the-art deep learning software tools.
\newblock In \emph{2016 7th International conference on cloud computing and big data (CCBD)}, pp.\  99--104. IEEE, 2016.

\bibitem[Singh et~al.(2025)Singh, Nan, Wang, D'Souza, Kapoor, {\"U}st{\"u}n, Koyejo, Deng, Longpre, Smith, et~al.]{singh2025leaderboard}
Singh, S., Nan, Y., Wang, A., D'Souza, D., Kapoor, S., {\"U}st{\"u}n, A., Koyejo, S., Deng, Y., Longpre, S., Smith, N.~A., et~al.
\newblock The leaderboard illusion.
\newblock \emph{arXiv preprint arXiv:2504.20879}, 2025.

\bibitem[Sutton et~al.(1998)Sutton, Barto, et~al.]{sutton1998reinforcement}
Sutton, R.~S., Barto, A.~G., et~al.
\newblock \emph{Reinforcement learning: An introduction}, volume~1.
\newblock MIT press Cambridge, 1998.

\bibitem[Thiyagalingam et~al.(2022)Thiyagalingam, Shankar, Fox, and Hey]{thiyagalingam2022scientific}
Thiyagalingam, J., Shankar, M., Fox, G., and Hey, T.
\newblock Scientific machine learning benchmarks.
\newblock \emph{Nature Reviews Physics}, 4\penalty0 (6):\penalty0 413--420, 2022.

\bibitem[Thrush et~al.(2022)Thrush, Tirumala, Gupta, Bartolo, Rodriguez, Kane, Rojas, Mattson, Williams, and Kiela]{thrush2022dynatask}
Thrush, T., Tirumala, K., Gupta, A., Bartolo, M., Rodriguez, P., Kane, T., Rojas, W.~G., Mattson, P., Williams, A., and Kiela, D.
\newblock Dynatask: A framework for creating dynamic ai benchmark tasks.
\newblock \emph{arXiv preprint arXiv:2204.01906}, 2022.

\bibitem[Waggoner(2025)]{waggoner2025}
Waggoner, P.
\newblock Democratizing algorithm development: Rethinking the design of complex hybrid decision-making systems.
\newblock In H.~Cherifi, L. M.~Rocha, C. C. Z.~E. (ed.), \emph{Proceedings of the 14th International Conference on Complex Networks and their Applications (COMPLEX NETWORKS 2025)}, Binghamton, NY, 2025. Springer Nature.

\bibitem[Wang et~al.(2018)Wang, Singh, Michael, Hill, Levy, and Bowman]{wang2018glue}
Wang, A., Singh, A., Michael, J., Hill, F., Levy, O., and Bowman, S.
\newblock Glue: A multi-task benchmark and analysis platform for natural language understanding.
\newblock In \emph{Proceedings of the 2018 EMNLP workshop BlackboxNLP: Analyzing and interpreting neural networks for NLP}, pp.\  353--355, 2018.

\bibitem[Wang et~al.(2024{\natexlab{a}})Wang, Hertzmann, and Russakovsky]{wang2024benchmark}
Wang, A., Hertzmann, A., and Russakovsky, O.
\newblock Benchmark suites instead of leaderboards for evaluating ai fairness.
\newblock \emph{Patterns}, 5\penalty0 (11), 2024{\natexlab{a}}.

\bibitem[Wang et~al.(2024{\natexlab{b}})Wang, Ma, Zhang, Ni, Chandra, Guo, Ren, Arulraj, He, Jiang, et~al.]{wang2024mmlu}
Wang, Y., Ma, X., Zhang, G., Ni, Y., Chandra, A., Guo, S., Ren, W., Arulraj, A., He, X., Jiang, Z., et~al.
\newblock Mmlu-pro: A more robust and challenging multi-task language understanding benchmark.
\newblock \emph{Advances in Neural Information Processing Systems}, 37:\penalty0 95266--95290, 2024{\natexlab{b}}.

\bibitem[Z{\"o}ller \& Huber(2021)Z{\"o}ller and Huber]{zoller2021benchmark}
Z{\"o}ller, M.-A. and Huber, M.~F.
\newblock Benchmark and survey of automated machine learning frameworks.
\newblock \emph{Journal of artificial intelligence research}, 70:\penalty0 409--472, 2021.

\end{thebibliography}
\bibliographystyle{icml2026}

%%%%%%%%%%%%%%%%%%%%%%%%%%%%%%%%%%%%%%%%%%%%%%%%%%%%%%%%%%%%%%%%%%%%%%%%%%%%%%%
%%%%%%%%%%%%%%%%%%%%%%%%%%%%%%%%%%%%%%%%%%%%%%%%%%%%%%%%%%%%%%%%%%%%%%%%%%%%%%%
% APPENDIX
%%%%%%%%%%%%%%%%%%%%%%%%%%%%%%%%%%%%%%%%%%%%%%%%%%%%%%%%%%%%%%%%%%%%%%%%%%%%%%%
%%%%%%%%%%%%%%%%%%%%%%%%%%%%%%%%%%%%%%%%%%%%%%%%%%%%%%%%%%%%%%%%%%%%%%%%%%%%%%%
%\newpage
%\appendix
%\onecolumn
%\section{You \emph{can} have an appendix here.}
%
%You can have as much text here as you want. The main body must be at most $8$
%pages long. For the final version, one more page can be added. If you want, you
%can use an appendix like this one.
%
%The $\mathtt{\backslash onecolumn}$ command above can be kept in place if you
%prefer a one-column appendix, or can be removed if you prefer a two-column
%appendix.  Apart from this possible change, the style (font size, spacing,
%margins, page numbering, etc.) should be kept the same as the main body.
%%%%%%%%%%%%%%%%%%%%%%%%%%%%%%%%%%%%%%%%%%%%%%%%%%%%%%%%%%%%%%%%%%%%%%%%%%%%%%%
%%%%%%%%%%%%%%%%%%%%%%%%%%%%%%%%%%%%%%%%%%%%%%%%%%%%%%%%%%%%%%%%%%%%%%%%%%%%%%%

\end{document}